%% file: main.tex
\documentclass[conference]{IEEEtran}
\usepackage[left=1.62cm,right=1.62cm,top=1.9cm, bottom=3.9cm]{geometry}
\IEEEoverridecommandlockouts
\usepackage{amsmath,amssymb,amsfonts}
\usepackage{pifont}
\usepackage[noend]{algorithmic}
\usepackage{algorithm}

\usepackage{subfigure}
\usepackage{graphicx}
\usepackage{textcomp}
\usepackage{lipsum}
\usepackage[backend=bibtex,style=ieee,maxcitenames=6,natbib=true]{biblatex} 
\usepackage{xcolor}
\def\BibTeX{{\rm B\kern-.05em{\sc i\kern-.025em b}\kern-.08em
    T\kern-.1667em\lower.7ex\hbox{E}\kern-.125emX}}

\usepackage{setspace}
\begin{document}

\title{Distributed Learning in Heterogeneous Environment: federated learning with adaptive aggregation and computation reduction}
\author{\parbox{6 in}{\centering Jingxin Li, Toktam Mahmoodi and Hak-Keung Lam \\
         Department of Engineering, 
        King's College London, London WC2R 2LS, U.K.\\
        E-mail:\{jingxin.1.li, toktam.mahmoodi, hak-keung.lam\}@kcl.ac.uk}
}

\maketitle
\thispagestyle{plain}
\pagestyle{plain}
\begin{abstract}
Although federated learning has achieved many breakthroughs recently, the heterogeneous nature of the learning environment greatly limits its performance and hinders its real-world applications. The heterogeneous data, time-varying wireless conditions and computing-limited devices are three main challenges, which often result in an unstable training process and degraded accuracy. Herein, we propose strategies to address these challenges. Targeting the heterogeneous data distribution, we propose a novel adaptive mixing aggregation (AMA) scheme that mixes the model updates from previous rounds with current rounds to avoid large model shifts and thus, maintain training stability. We further propose a novel staleness-based weighting scheme for the asynchronous model updates caused by the dynamic wireless environment. Lastly, we propose a novel CPU-friendly computation-reduction scheme based on transfer learning by sharing the feature extractor (FES) and letting the computing-limited devices update only the classifier. The simulation results show that the proposed framework outperforms existing state-of-the-art solutions and increases the test accuracy, and training stability by up to $\textbf{2.38}\%, \textbf{93.10}\%$ respectively. Additionally, the proposed framework can tolerate communication delay of up to 15 rounds under a moderate delay environment without significant accuracy degradation. 
\end{abstract}

\input{Section1}
\input{Section3} 
\input{Section5} 
\input{Section4} 
\input{Section6} 
\input{Section7} 


\renewcommand{\bibfont}{\footnotesize}
\begin{spacing}{0.95}
\printbibliography 
\end{spacing}

\end{document}

%% file: Section1.tex
\section{Introduction}\label{Sec1}
Recent years have witnessed the rise of a new distributed learning paradigm, Federated Learning (FL) \cite{McMahanMRHA17}, which protects user privacy by allowing remote clients to train the model collaboratively without sharing privacy-sensitive data. It also helps to save communication bandwidth by eliminating the need to transmit large data samples. Although FL has achieved many breakthroughs, the deployment of FL in practical applications still faces many challenges. The heterogeneous learning environment is one of the most challenging problems and it is mainly embodied in two aspects. The first one is the statistical heterogeneity, \textit{a.k.a.} the non-iid (independent identically distributed) data distribution. It is known that learning from a balanced dataset, where the samples are uniformly distributed across classes, plays a crucial role in boosting the model performance. However, data collected from geographically-distributed clients are often highly non-iid, which deteriorates the FL training performance, such as the accuracy and learning stability. On the other hand, system heterogeneity is the second one. Due to the distributed nature, each client has a different level of resources, such as computation capability, and communication bandwidth. Clients with limited computation resources can not finish the local training in time due to the limitation of the computing unit while dynamic wireless conditions may delay the transmission of the model updates sent by the clients. Under both scenarios, the server fails to receive the updates in time. Thus, the FL global aggregation is either postponed, which leads to long idling time for other clients, or conducted without these delayed updates, which may lead to lower accuracy and reduced stability. The problem becomes even worse in the non-iid environment. Therefore, tackling the heterogeneity issue is vital for promoting FL in real-world applications.

In fact, the FL heterogeneity issue has drawn much attention. Federated Averaging (FedAvg) \cite{McMahanMRHA17}, the naive FL optimisation method, claims to have tolerance of the non-iid data distribution. However, \cite{abs-1806-00582} proved that if the data distribution is extremely skewed, where each client only possesses the samples from one or two class(es), FedAvg fails to retain the model performance, and the accuracy is dropped by up to 11.31$\%$ for MNIST classification \cite{lecun1998gradient} and up to 51.31$\%$ for CIFAR10~\cite{krizhevsky2009learning}. Also, FedAvg does not address the system heterogeneity issue. Hence, delayed updates are commonly discarded, resulting in information loss and consequently leading to unstable, degraded training performance. However, under the non-iid setting, discarding such updates may result in converging to the local model optimum, \textit{a.k.a.} the client drift phenomenon~\cite{kumar2021coding}, and thus, such delayed model updates should be handled with careful consideration. Work in \cite{abs-1806-00582} demonstrated that the degraded performance under the non-iid data can be accounted for by the weight divergence and they also proposed to tackle the non-iid issue by sharing a small public dataset, which has a uniform distribution over classes. However, sharing the dataset between clients and the server violates the privacy rule of FL. Targeting the system heterogeneity, work in \cite{RahmanTMT21} and \cite{NishioY19} select participants with sufficient resources. However, this may lead to the \textit{fairness} issue, \textit{i.e.,} regardless of the selection criteria, \cite{RahmanTMT21} and \cite{NishioY19} both favour a specific group of clients, meaning the resource-limited clients have less chance of participating in the training process. Then it is unavoidable that these resource-limited clients would have lower accuracy on the trained model, especially in non-iid settings. FedProx \cite{LiSZSTS20} is another method addressing the heterogeneity issue in FL. For the statistical heterogeneity, FedProx proposes to add a proximal term in the local loss function to restrict the model updates. Regarding the system heterogeneity, FedProx allows the computing-limited devices to conduct fewer rounds of local updates depending on the computing capability. However, the delayed updates caused by the dynamic communication environment are not considered. 

Although the aforementioned methods achieved great success in optimising the FL performance under the heterogeneous environment, all the proposed frameworks focus only on the unilateral side of the heterogeneity issue, either the statistical or the system heterogeneity. To our knowledge, research effort on systematically addressing the FL heterogeneity is still scarce. Hence, to fill in the blank, we propose a framework to thoroughly demonstrate the heterogeneities in FL, including the non-iid data distribution, the computing-limited devices and the asynchronous model updates caused by the dynamic wireless conditions. We propose solutions to tackle these challenges and aim to restore training accuracy and stability. Contributions of the work are stated as follows:
\begin{itemize}
    \item We propose to share feature extractor (FES) among computing-limited devices to reduce the local computation burden. The CPU-friendly FES scheme is highly compatible with FL optimisation methods and can be easily implemented at large scales. To the best of our knowledge, we are the first work utilising the concept of feature extractor to mitigate the computation burden in FL;
    \item Targeting the non-iid data, we propose a novel adaptive mixing aggregation (AMA) scheme that mixes the current model updates with the previous one using an adaptive weighting scheme, to improve the training accuracy and stability. Previously in FL, such weighted aggregation schemes are often used to handle the asynchronous updates. To the best of our knowledge, we are the first ones formally analysing the effect of mixing aggregation on statistical heterogeneity. Furthermore, to address the wireless dynamics, we propose to incorporate the delayed model updates with a novel staleness-based weighting function in the AMA scheme;
    \item Simulation results show that the proposed AMA-FES FL framework increases the accuracy by up to 19.77$\%$ and 2.38$\%$ on FMNIST \cite{xiao2017/online} when compared with the naive FL and FedProx respectively, and up to 1.257$\%$ on MNIST compared with FedProx. Under a moderate delay scenario, it shows that the asynchronous aggregation scheme allows a maximum delay of up to 15 communication rounds without loss of accuracy.
\end{itemize}

%% file: Section3.tex
\section{Learning Model}\label{Sec3}
\input{FIG}
Herein, we consider the training problem of FL on classification tasks in a dynamic wireless environment, as shown in Fig~\ref{fig:AMAFES FL}. Each client trains the model locally and then uploads the model updates to the server for global aggregation. We consider a set of clients, $\{\mu_i, i\in \mathcal{K}\}$, where $\mathcal{K}=\{1,2,...,K\}$ denotes the client index, collaboratively train the task model in $B$ communication rounds\footnote{In this work, we use the term communication rounds and training rounds interchangeably.}. At the start of each training round, a subset of clients, $\{\mu_i, i \in k_t\}$, where $k_t \subseteq \mathcal{K}$, are randomly selected for training. $k_t$ denotes the indexes of the selected clients in the training round $t$ and $|k_t|=m$ denotes the size of the selected clients. Each client $\mu_i, \forall i\in \mathcal{K}$ owns a local dataset $d_i$, with a size of $|d_i|>0$. If a client $\mu_i$ is selected in training round $t$, then the distributed global model $\omega^{t-1}$ will be trained on the local dataset $d_i$ for $e$ local epochs. Note that the global model is updated at the end of each training round, which means at the start of training round $t$, $\omega^{t-1}$ is distributed to clients for local training, as shown in Fig.~\ref{fig:AMAFES FL}.

We consider training with deep neural network (DNN) for the classification task, where the clients collaboratively train the DNN model to minimise the following loss function,
\begin{equation}
   \min_{\omega} \sum_{i=1}^{K} \frac{|d_i|}{|D|} F_i (\omega),\  F_i (\omega) = \frac{1}{|d_i|} \sum_{(x_j,y_j) \in d_i} f_j(\omega).
\end{equation}
$|D|$ denotes the size of the dataset owned by all the clients. $f_j(\omega)$ represents the local loss function $\ell (x_j,y_j; \omega)$ on example $(x_j,y_j) \in d_i$ given model parameter $\omega$. Herein, we use cross-entropy \cite{de2005tutorial} as the local loss function for the training task.


%% file: FIG.tex
\begin{figure}[t]
    \centering
    \includegraphics[width=239pt]{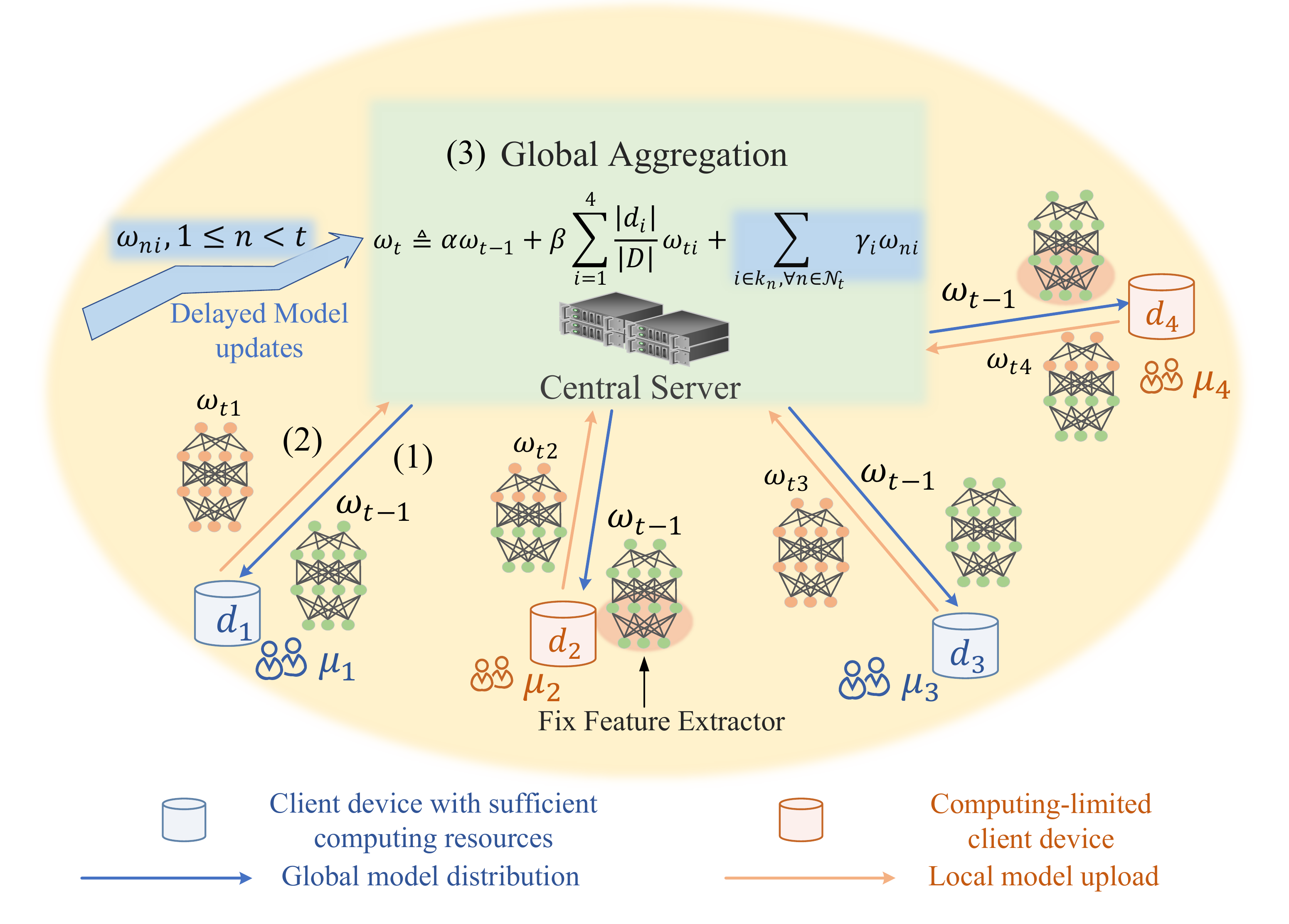}
    \caption{Demonstration of the proposed AMA-FES FL framework at communication round $t$, (1) (2) (3) indicates the chronological order of each step.}
    \label{fig:AMAFES FL}
\end{figure}

%% file: Section5.tex
\section{Feature Extractor Sharing (FES) -based computation reduction scheme}\label{Sec5}
\input{Algorithm1}
Machine learning models are often large and may require high-profile processing units to achieve the desired performance. In FL, the computing power of the client devices greatly varies and those with low computing power can significantly delay the model training. Thus, the question here is how to proceed with the training on computing-limited devices without affecting the accuracy and efficiency of the training process. Unlike most model compression schemes that target the model inference, the research work in \cite{abs-2204-12703} proposes an ensemble FL algorithm, FED-ET, to provide computation reduction based on knowledge transfer. A set of small models, differing in structure, are pre-defined to accommodate computation heterogeneities of client devices.  After being trained on clients, these small models are used to train the global model at the server via a weighted distillation scheme. Although experiments prove that FED-ET is robust to both data and device heterogeneities, it can not be implemented at a large scale since a variety of small models need to be defined beforehand. In contrast, FedProx directly reduces the computation during the local training by allowing computing-limited devices to compute fewer rounds. However, the modern neural model often contains convolutional layers, which require intensive computation and specialised hardware, such as GPU, to speed up the calculation. Therefore, despite reducing the computation, training with FedProx takes longer for devices without GPU and thus, it is not CPU-friendly. Hence, an ideal way of mitigating the computation burden and also speeding up the local training for these computing-limited devices would be to avoid training these convolutional layers. 

Inspired by the concept of feature-extractor in transfer learning, we propose to mitigate the computation burden for computing-limited devices by letting them update only the classifier while the devices with sufficient computing resources train the whole model. The classifier in neural models often consists of fully-connected (FC) layers and requires less computation compared to the feature extractor, which consists of several convolutional layers. Earlier work in \cite{LuoCHZLF21} conducts a thorough analysis of the similarities between layers across clients in the non-iid setting and observes that the parameters of the classifier have the lowest similarities while earlier layers exhibit higher conformity. This finding justifies the rationality of training only the classifier under computation insufficiency. 

Suppose at the start of a communication round $t$, the distributed global model $\omega_{t-1}$ consists of a feature extractor $\omega_{t-1}^{f}$ and a classifier $\omega_{t-1}^{c}$, as shown in \eqref{eq:10}. During the local training session, clients with computing-limited devices fix weights for $\omega_{t-1}^{f}$ and only update the classifier $\omega _{t-1}^{c}$, as shown in \eqref{eq:11}, where $\epsilon$ is the learning rate, $\omega_{ti}^{c}$ is the updated classifier of a client $\mu_i$ and $\omega_{ti}$ denotes the updated local model for $\mu_i$. 
\begin{equation}\label{eq:10}
    \omega_{t-1} = \left[ \begin{array}{l}
      \omega_{t-1}^{f}\\
      \omega_{t-1}^{c} 
    \end{array}  \right], 
\end{equation}
\begin{equation}\label{eq:11}
    \omega_{ti}^{c}  \longleftarrow \omega_{t-1}^{c}  - \epsilon \nabla_{\omega} \mathbf{\ell}(d_i; \omega _{t-1}^{c}), \ \omega_{ti} = \left[ \begin{array}{l}
      \omega_{t-1}^{f}\\
      \omega_{ti}^{c} 
    \end{array}  \right].
\end{equation}
The FES scheme can be treated as a complementary scheme to the partial work scheme in FedProx. FES avoids training the convolutional layers and thus is CPU-friendly. It provides more flexibility in lowering the computation cost and only requires minor modification of the algorithm. Thus, FES is highly compatible with other FL optimisation methods and, unlike FED-ET, it can easily be implemented at large scales.

%% file: Algorithm1.tex
\begin{algorithm}[t]
\caption{AMA-FES FL Framework}\label{alg:Sec4_1}
\begin{algorithmic}[1]
\STATE{\bfseries{Initialise:}} $e, B, \mathcal{K}, p, m, \alpha_0, \gamma, \eta, b, \omega$
\STATE{\bfseries{Server Executes:}}
\FOR{$t=1$ {\bfseries to} B}
\STATE Select a subset of clients, $k_t \subseteq \{1,2,3,...,K\}, |k_t|=m$

\FOR{$\mu_i \in k_t$ in parallel}
\STATE Client $\mu_i$ updates local model
\ENDFOR

\IF{asynchronous model updates received}
\STATE Update the global model $\omega_t$ $\to$ \COMMENT{Asynchronous AMA scheme in \ref{Sec4Sub2}} 
\ELSE
\STATE Update the global model $\omega_t$ $\to$\COMMENT{AMA scheme in \ref{Sec4Sub1}}
\ENDIF

\ENDFOR

\STATE{\bfseries{Client Executes:}}
\IF{client $\mu_i$ is computing-limited}
\STATE Update the model classifier $\omega^f$only $\to$ \COMMENT{Eq. \eqref{eq:11}} 
\ELSIF{client $\mu_i$ has sufficient computing resources}
\STATE Update the whole model $\omega$  
\ENDIF
\STATE Upload the trained local model $\omega_{ti}$ to the server
\end{algorithmic}
\end{algorithm}

%% file: Section4.tex
\section{Adaptive Mixing Aggregation (AMA) Scheme}\label{Sec4}
After receiving the local model updates, the server conducts model aggregation to update the global model. In this section, we introduce a novel AMA scheme to mitigate the impacts of non-iid data and asynchronous model updates caused by wireless dynamics. 
\subsection{AMA for Statistical Heterogeneity}\label{Sec4Sub1}
Under the non-iid environment, each client has samples from only a few classes, upon which the local model is trained. Consequently, the local model tends to \textit{overfit} these classes, resulting in lower accuracy on samples from the remaining classes. Similarly, in an FL communication round, the selected clients may only have samples from several classes, \textit{i.e.} low diversity in the training dataset. Then, the aggregated model would favour samples from these classes and generalise poorly on the balanced dataset. Since the clients are randomly selected, the diversity of the training data is uncertain, leading to unstable training performance. Therefore, the problem can be mitigated by either increasing the diversity in $D$, \textit{i.e.} sharing public dataset \cite{abs-1806-00582} \cite{ZhuHZ21}, which violates the privacy rule of FL, or penalising the model updates in each communication round, which avoids large changes in the model updates from round to round and thus ensures stable training. As shown in \eqref{eq:1}, existing solutions, for example, FedProx \cite{LiSZSTS20}, restrict the model updates by adding a regularisation term upon the initial model weights $\omega_0$ and the current model weights $\omega$ to the local loss function $\ell (x,y; \omega)$, 
\begin{equation}\label{eq:1}
    f(\omega) \overset{\Delta}{=} \ell (x_j,y_j; \omega) + \rho || \omega - \omega_0 ||^2,
\end{equation}
where $\rho$ is a user-specified hyperparameter that controls the regularisation strength. However, $\rho$ is sensitive and requires careful consideration. Small $\rho$ does not provide enough regularisation strength while large $\rho$ hurts the model convergence. 
 
In contrast, we propose a novel adaptive mixing aggregation (AMA) scheme to restore training accuracy and stability. Instead of restricting the local model updates, we directly take the weighted average of the distributed global model $\omega_{t-1}$ and the received model updates $\omega_{ti}$, as shown in \eqref{eq:2},
\begin{equation}\label{eq:2}
    \omega_t \overset{\Delta}{=} \alpha\  \omega_{t-1} + \beta \sum_{i \in k_t} \frac{|d_i|}{|D|}\  \omega_{ti},\  \beta = 1 - \alpha.
\end{equation}
Specifically, $\omega_{ti}$ denotes the model updates from client $\mu_i$ in communication round $t$. $\alpha$ and $\beta$ control how much the new global model $\omega_t$ learns from $\omega_{t-1}$ and $\{\omega_{ti}, \forall{i}\in k_t \}$ respectively. By carefully and gradually learning new knowledge from the client-side models and avoiding sudden shifts to a new model, which may be greatly biased towards certain classes, the training stability is increased. Nevertheless, if $\alpha$ is large, we are not learning much new knowledge, leading to slow convergence while small $\alpha$, $i.e.,$ large $\beta$, may not warrant training stability. Hence, to balance the convergence rate and the stability, instead of fixing $\alpha$ and $\beta$ to constant numbers, we set $\alpha = \alpha_0 + \eta t,$ where $\alpha_0$ is the initial value for $\alpha$, $\eta$ is the increase rate and $t$ is the training round index. At the start of the training, $\alpha$ is small and $\beta$ is respectively large, allowing fast convergence. As the training continues, $\alpha$ increases and $\beta$ becomes smaller, which gradually increases the stability while keeping learning new knowledge. Experiments in section \ref{Sec6} show that this adaptive mixing scheme achieved better accuracy and training stability than the regularisation scheme, FedProx.    

\textit{Remark 1}: The work in \cite{9378161} introduces a similar balancing scheme. However, it targets the local training stage while our method targets global aggregation. Besides, the local model balancing scheme in \cite{9378161} aim to address the online learning issue while our AMA scheme targets non-iid data.
\subsection{Asynchronous AMA for delayed model updates}\label{Sec4Sub2}
Instability, such as variation in data rate, is an intrinsic nature of wireless communication. Even with careful resource scheduling, factors such as the weather, and unexpected moving obstacles, can greatly affect the channel condition and thus, delay the model updates. In the FL training, the server can not afford to wait for all the delayed updates before conducting the aggregation since it would result in an excessively long training time. Additionally, discarding such delayed updates results in information loss and reduced accuracy. Therefore, most existing approaches incorporate asynchronous updates into the global model with sophisticated weighting schemes. In this work, following this convention, we propose to add an asynchronous term to the AMA with a novel weighting scheme to account for the delayed updates. Notably, unlike the fully asynchronous scheme in \cite{xie2019asynchronous}, which leads to large communication overheads, we deploy the periodically asynchronous scheme, meaning that the received asynchronous updates would only be aggregated at the end of each training round. 
\begin{equation}\label{eq:5}
    \omega_t \overset{\Delta}{=} \alpha\  \omega_{t-1} + \beta \sum_{i \in k_t} \frac{|d_i|}{|D|}\  \omega_{ti} + \sum_{i \in k_{n}, \forall n  \in \mathcal{N}_t} \gamma_i \omega_{ni}
\end{equation}
Equation \eqref{eq:5} shows the details of the asynchronous aggregation scheme. $k_{n}$ indicates the set of clients, whose updates were delayed from training round $n, 1\leq n < t$ and received by the server in the current training round $t$. $\mathcal{N}_t$ denotes the time index of all delayed model updates received at training round $t$. $\gamma_i$ represents the staleness-based weight for the delayed model updates of client $\mu_i$. Specifically, we set,
\begin{equation}\label{eq:6}
    \alpha + \beta  + \sum_{i \in k_{n}, \forall n  \in \mathcal{N}_t} \gamma_i = 1.
\end{equation}
Both the $\alpha$-term and the $\gamma$-term in equation \eqref{eq:5} represent model weights from previous rounds. Hence, following settings in \ref{Sec4Sub1}, we set 
\begin{equation}\label{eq:7}
    \alpha + \sum_{i \in k_{n},\forall n  \in \mathcal{N}_t} \gamma_i = \alpha_0 + \eta t,
\end{equation}
such that the weight for the client-side model updates, $\beta = 1 - (\alpha_0 + \eta t)$, remains the same as in section \ref{Sec4Sub1}.
\begin{figure*}[t]
    \centering
    \includegraphics[height=185pt]{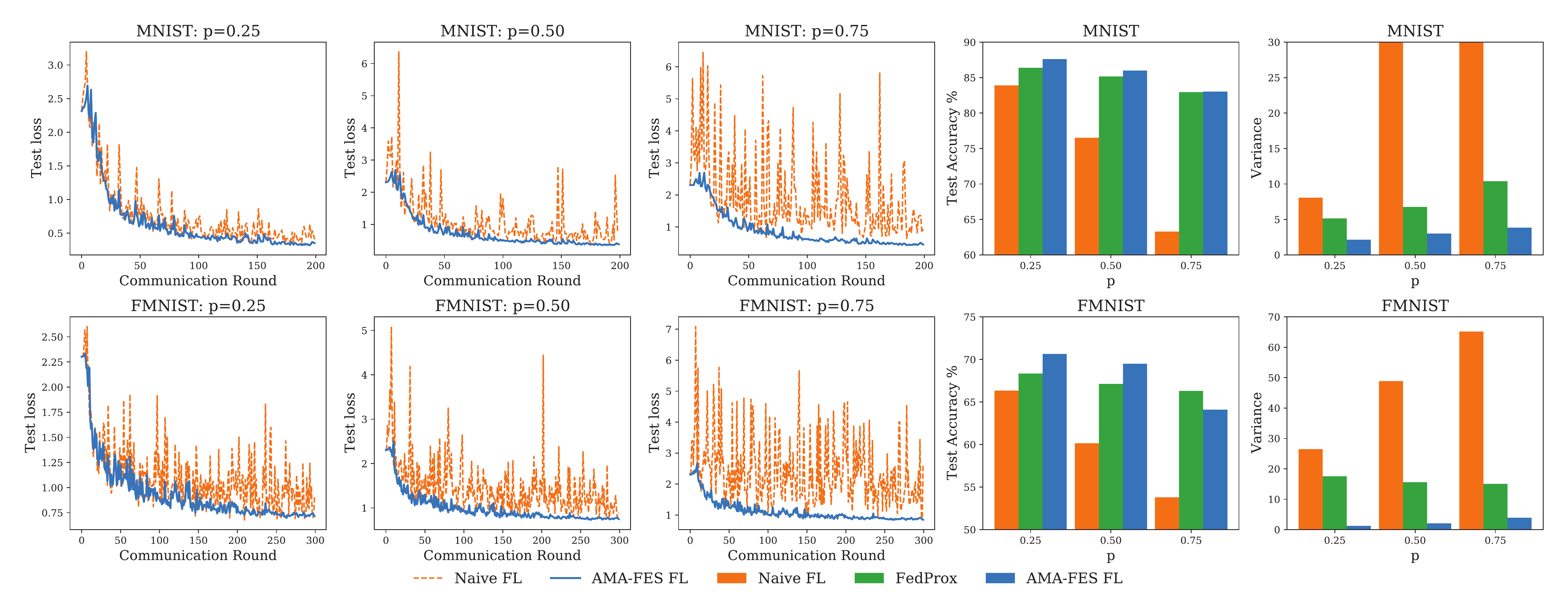}

    \caption{(Left) Convergence performance of AMA-FES FL compared with the naive FL; (Right) Testing accuracy and training stability: for p=0.50 and 0.75, the variance of the naive FL on MNIST is 86.96 and 173.16 respectively. For ease of visualisation, the scale of the y-axis is cut to [0, 30].} 
    \label{fig:Sync Convergence}
\end{figure*}
Now, we need to pick the function of $\gamma$ based on the staleness of the delayed updates. The $\alpha$-term controls the model updates from the previous training round, which has a staleness no larger than the $\gamma$-terms, \textit{i.e.} $t-(t-1) \leq t-n$. Hence, $\alpha$ should be the dominator and has a larger value than the $\gamma$-terms. After conducting several experiments, we found the following setting gave the best performance,
\begin{equation}\label{eq:9}
    \gamma_{i\_} = b(1-\sigma(t-n)), \ \ \alpha_{\_}=1-\sigma(1),
\end{equation}
and we normalise each term, 
\begin{equation}
    \alpha = \frac{\alpha_{\_}}{\alpha_{\_}+ \sum_{i \in k_{n}, \forall n  \in \mathcal{N}_t} \gamma_{i\_}} (\alpha_0 + \eta t),
\end{equation}
\begin{equation}
     \gamma_{i} = \frac{ \gamma_{i\_}}{\alpha_{\_}+ \sum_{i \in k_{n}, \forall n  \in \mathcal{N}_t} \gamma_{i\_}} (\alpha_0 + \eta t).
\end{equation}
 $b$ is a constant that controls the strength of the $\gamma$-term with a value range preferably from 0.2 to 1, $(t-n)$ is the staleness of the model updates for client $\mu_i$, and $\sigma(\cdot)$ represents the standard sigmoid function.

%% file: Section6.tex
\section{Simulation Results}\label{Sec6}
\begin{table}[t]
    \centering
    \footnotesize
    \caption{Simulation parameters}
    \label{tab:para setting}
    \begin{tabular}{|c|c|}
    \hline
         \textbf{Parameter} & \textbf{Value} \\
         \hline 
        AMA parameters, $\alpha_0,\eta, b$ & 0.1, 2.5$e^{-3}$, 0.6 \\
        FedProx regularisation strength, $\rho$ & 0.01\\
        Total number of clients,  $K$ & 50\\
        $\#$ of clients selected in each training round, $m$ & 10\\
         Training rounds, $B$ for MNIST, FMNIST & 200, 300\\
          Local epochs, $e$ & 10 \\
          Ratio of computing-limited devices, p & 0.25, 0.50, 0.75\\
          Max. delay for async. model updates & 5, 10, 15 rounds\\
          Prob. transmission delay (Moderate, Severe) & 30$\%$, 70$\%$ \\
          Learning rate, $\epsilon$ & 0.001\\
          \hline
    \end{tabular}
\end{table}
In this section, we investigate the performance of the AMA-FES FL framework, which is summarised in \textbf{Algorithm \ref{alg:Sec4_1}}, under the heterogeneous environment with classification tasks on MNIST \cite{lecun1998gradient} and FMNIST \cite{xiao2017/online} images. We consider a small DNN with 2 convolutional layers, of which the kernel size is $5\times5$, and with 3 FC layers. We use cross-entropy as the local loss function. Specifically, for FL, we set the total communication rounds $B$ for MNIST and FMNIST to be 200, and 300 respectively, with local epochs $e$ as 10 for both datasets. We run the experiments over $K=50$ clients and during each communication round, $m=10$ clients are randomly selected to perform local training. The key performance indicators of the experiments are test accuracy and training stability. Specifically, the training stability is measured using the variance of the converged test accuracy for the last 50 training rounds. Hence, the larger the variance, the lower the training stability, and \textit{vice versa}. 

To impose the data heterogeneities, we follow the non-iid settings in \cite{McMahanMRHA17}, where each client only has access to samples from two classes. To simulate the computation heterogeneities, we set the ratio of computing-limited devices that can not provide timely updates, p, as 0.25, 0.50 and 0.75 in three different runs of the simulation. Regarding the computation reduction scheme, all the computing-limited devices are assumed the same and train only the final three FC layers. Regarding the wireless dynamics, we set the probability of clients experiencing transmission delay to $30\%$ and $70\%$, to imitate moderate and severe communication delay environments, where the model updates are received by the server in later training rounds. Specifically, we experiment with a maximum transmission delay of 5, 10 and 15 training rounds in each communication environment. We compare the training performance of the proposed AMA-FES FL with the naive FL, which directly drops the computing-limited devices from training and conducts the global aggregation without any mixing scheme. We also use FedProx as a benchmark scheme. Table \ref{tab:para setting} shows the detailed parameter settings for the simulations. 

\subsection{AMA-FES FL for synchronous model updates}\label{sub1}
Fig. \ref{fig:Sync Convergence} (left-side line charts) show the convergence performance of the AMA-FES FL framework on MNIST and FMNIST under different levels of computation heterogeneities. From the left-side line charts, it shows that the test loss of the naive FL (orange dash line) becomes more fluctuated as p increases. In contrast, the proposed AMA-FES FL scheme (blue solid line) manages to smooth the curve and converge to a much smaller loss value for both datasets. The right-side bar charts show the test accuracy and the training stability with different levels of computation heterogeneities. It shows that AMA-FES FL has better accuracy than both the naive FL and the FedProx scheme in most settings, except that when p$=0.75$, FedProx has slightly higher accuracy on FMNIST. A possible explanation is that the number of computing-limited devices is large and the remaining devices, which train the whole model, failed to provide a decent feature extractor. In addition, the proposed scheme has a much lower variance than the naive FL and FedProx, which means higher stability. Specifically,  the AMA-FES FL framework increases the accuracy by at most \textbf{19.77$\%$} on MNIST when compared with the naive FL and \textbf{2.38$\%$} on FMNIST when compared with FedProx. Furthermore, the AMA-FES FL increases the training stability by up to \textbf{93.10$\%$} on FMNIST, compared with FedProx. 


\subsection{AMA-FES FL with asynchronous model updates}\label{sub2}
\begin{figure}[t]
    \centering

    \includegraphics[width=230pt]{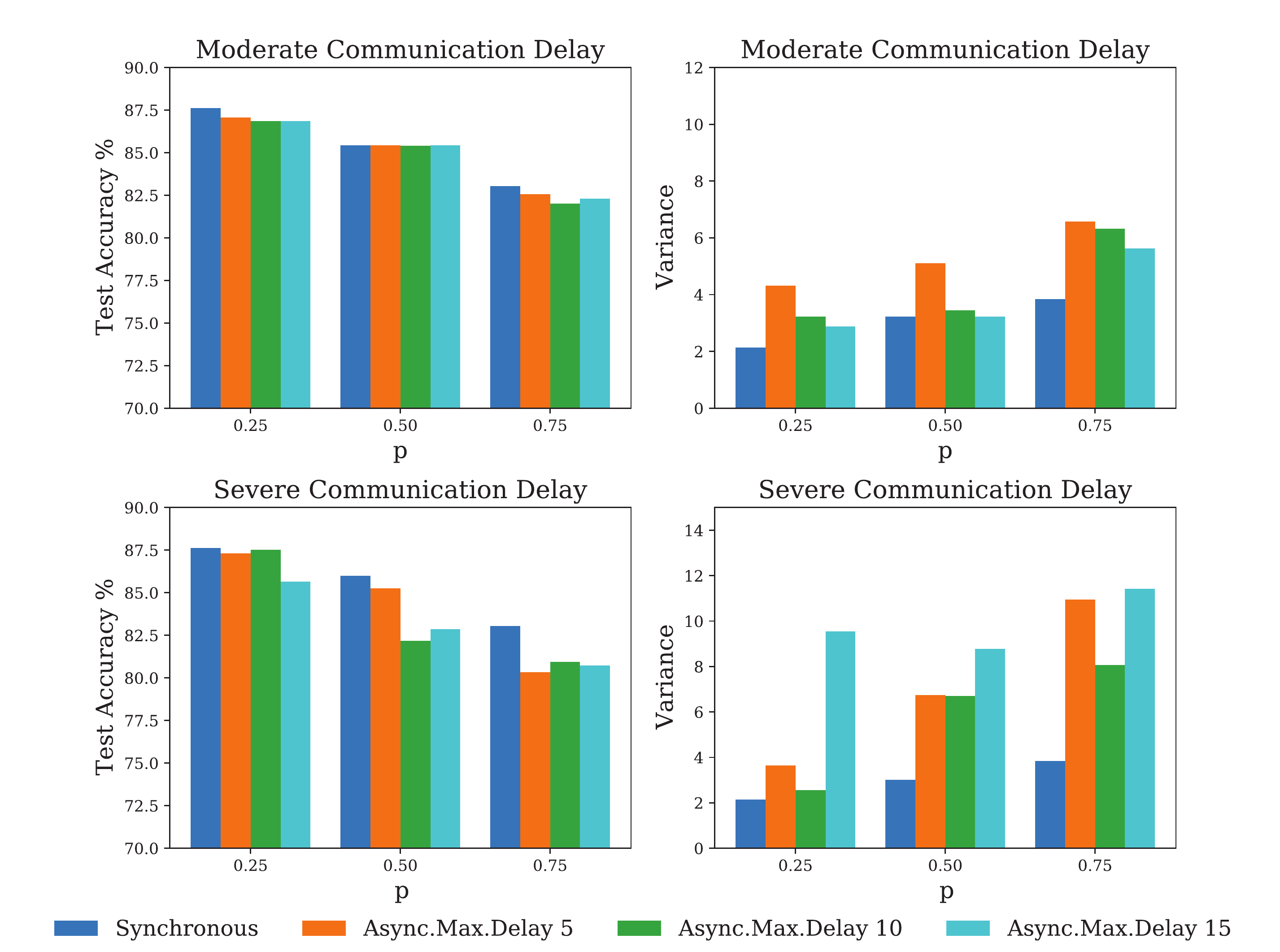}
    \caption{Test accuracy and training stability statistics in (up) Moderate communication delay environment and (bottom) Severe communication delay environment on MNIST classification with different levels of $p$.}
    \label{fig:AsyncFigs}
\end{figure}
Fig. \ref{fig:AsyncFigs} shows the test accuracy and training stability of the asynchronous AMA scheme under moderate and severe communication delay environments on MNIST image classifications. In the moderate delay communication environment, as the level of maximum delay increases from 5 rounds to 15 rounds, we can observe that the accuracy degradation is quite small, less than 1$\%$ for all settings, with acceptable stability reductions. On the other hand, under the severe communication delay environment, the accuracy fluctuation becomes relatively larger, up to $2.72\%$, as more clients experience transmission delay. The training stability reduction also becomes more noticeable, compared with the moderate communication environment. However, under the severe delay environment, the proposed scheme still manages to maintain accuracy with minor training stability reduction for a maximum delay of up to 10 rounds when p$=0.25$. 



%% file: Section7.tex
\section{Conclusion}\label{Sec7}
This work studies the problem of environmental heterogeneity in federated learning. Tackling the statistical heterogeneity, we propose an adaptive mixing aggregation scheme that combines the model updates from the previous and current rounds to avoid large model weight shifts caused by the skewed data distribution. We also propose an asynchronous AMA scheme to account for model updates delayed by the dynamic wireless channels with a novel weighting scheme based on the model staleness. Finally, targeting the computing-limited devices, we alleviate the computation burden by fixing the feature extractor and letting these devices train only the classifier. Simulation results reveal that the proposed framework significantly improves the training accuracy and stability and outperforms the state-of-the-art method FedProx in most scenarios. It also shows that the asynchronous AMA scheme is highly robust to delayed model updates under the moderate delay environment. 

\footnotesize
\section*{Acknowledgements}
This work has received funding from Innovate UK in the project ANIARA: Automation of Network edge Infrastructure and Applications with aRtificiAl intelligence, CELTIC-NEXT under grant agreement.

%% file: main.bib
@article{kumar2021coding,
  title={Coding for Straggler Mitigation in Federated Learning},
  author={Kumar, Siddhartha and Schlegel, Reent and Rosnes, Eirik and others},
  journal={arXiv preprint arXiv:2109.15226},
  year={2021}
}

@inproceedings{McMahanMRHA17,
  author    = {Brendan McMahan and
               Eider Moore and
               Daniel Ramage and
               Seth Hampson and
               Blaise Ag{\"{u}}era y Arcas},

  title     = {Communication-Efficient Learning of Deep Networks from Decentralized
               Data},
   booktitle={Proc. 20th Int. Conf. Artif. Intell. Stat.},
  volume    = {54},
  pages     = {1273--1282},
  year      = {April 2017}
}

@article{abs-1806-00582,
  author    = {Yue Zhao and
               Meng Li and
               Liangzhen Lai and
               Naveen Suda and
               Damon Civin and
               Vikas Chandra},
  title     = {Federated Learning with Non-IID Data},
  journal   = {CoRR},
  volume    = {abs/1806.00582},
  year      = {2018},
  eprinttype = {arXiv},
  eprint    = {1806.00582}
}

@inproceedings{LiSZSTS20,
 title={Federated optimization in heterogeneous networks},
  author={Li, Tian and Sahu, Anit Kumar and Zaheer, Manzil and Sanjabi, Maziar and Talwalkar, Ameet and Smith, Virginia},
  journal={Proceedings of Machine Learning and Systems},
  volume={2},
  pages={429--450},
  year={2020}
}

@inproceedings{LuoCHZLF21,
   title={No fear of heterogeneity: Classifier calibration for federated learning with non-iid data},
  author={Luo, Mi and Chen, Fei and Hu, Dapeng and Zhang, Yifan and Liang, Jian and Feng, Jiashi},
  journal={Advances in Neural Information Processing Systems},
  volume={34},
  pages={5972--5984},
  year={2021}
}

@inproceedings{ZhuHZ21,
   title={Data-free knowledge distillation for heterogeneous federated learning},
  author={Zhu, Zhuangdi and Hong, Junyuan and Zhou, Jiayu},
  booktitle={ICML},
  pages={12878--12889},
  year={2021}
}

@article{abs-2204-12703,
   title={Heterogeneous Ensemble Knowledge Transfer for Training Large Models in Federated Learning},
  author={Cho, Yae Jee and Manoel, Andre and Joshi, Gauri and Sim, Robert and Dimitriadis, Dimitrios},
  journal={arXiv preprint arXiv:2204.12703},
  year={2022}
}

@article{RahmanTMT21,
  author    = {Sawsan Abdul Rahman and
               Hanine Tout and
               Azzam Mourad and
               Chamseddine Talhi},
  title     = {\text{FedMCCS}: Multicriteria Client Selection Model for Optimal \text{IoT} Federated Learning},
  journal   = {{IEEE} Internet Things J.},
  volume    = {8},
  number    = {6},
  pages     = {4723--4735},
  year      = {2021}
}

@inproceedings{NishioY19,
  title={Client selection for federated learning with heterogeneous resources in mobile edge},
  author={Nishio, Takayuki and Yonetani, Ryo},
  booktitle={Proc. IEEE Int. Conf. Commun.(ICC)},
  pages={1--7},
  year={2019}
}

@article{xie2019asynchronous,
  title={Asynchronous federated optimization},
  author={Xie, Cong and Koyejo, Sanmi and Gupta, Indranil},
  journal={arXiv preprint arXiv:1903.03934},
  year={2019}
}

@article{lecun1998gradient,
  title={Gradient-based learning applied to document recognition},
  author={LeCun, Yann and Bottou, L{\'e}on and Bengio, Yoshua and Haffner, Patrick},
  journal={Proceedings of the IEEE},
  volume={86},
  number={11},
  pages={2278--2324},
  year={1998},
  publisher={Ieee}
}

@online{xiao2017/online,
  author       = {Han Xiao and Kashif Rasul and Roland Vollgraf},
  title        = {Fashion-\text{MNIST}: a Novel Image Dataset for Benchmarking Machine Learning Algorithms},
  year         = {2017}
}

@article{de2005tutorial,
  title={A tutorial on the cross-entropy method},
  author={De Boer, Pieter-Tjerk and Kroese, Dirk P and Mannor, Shie and Rubinstein, Reuven Y},
  journal={Annals of operations research},
  volume={134},
  number={1},
  pages={19--67},
  year={2005},
  publisher={Springer}
}

@article{krizhevsky2009learning,
  title={Learning multiple layers of features from tiny images},
  author={Krizhevsky, Alex and Hinton, Geoffrey and others},
  year={2009},
  publisher={Citeseer}
}

@INPROCEEDINGS{9378161,  author={Chen, Yujing and Ning, Yue and Slawski, Martin and Rangwala, Huzefa},  booktitle={IEEE International Conference on Big Data},   title={Asynchronous Online Federated Learning for Edge Devices with Non-IID Data},   year={2020},   pages={15-24}  }
